\documentclass[conference]{IEEEtran}
\IEEEoverridecommandlockouts
\usepackage{caption}
\usepackage{cite}
\usepackage{textcomp}
\usepackage{graphicx}
\usepackage{multirow}
\usepackage{amsmath,amssymb,amsfonts}
\usepackage{booktabs}
\usepackage{url}
\usepackage{xcolor}
\usepackage{algorithm}
\usepackage{algpseudocode}
\usepackage{algorithmicx}
\usepackage{subcaption}
\usepackage{listings}
\usepackage{manyfoot}
\usepackage{amsthm}
\usepackage{mathrsfs}
\usepackage{orcidlink}
\def\BibTeX{{\rm B\kern-.05em{\sc i\kern-.025em b}\kern-.08em
T\kern-.1667em\lower.7ex\hbox{E}\kern-.125emX}}

\title{Machine Learning Algorithms for Detecting Mental Stress in College Students}

\author{
  \IEEEauthorblockN{Ashutosh Singh\textsuperscript{\textdagger}\orcidlink{0009-0007-7210-0536}, Khushdeep Singh\orcidlink{0009-0009-9171-7632}, Amit Kumar\orcidlink{0009-0009-4973-0358}, Abhishek Shrivastava \orcidlink{0009-0007-7101-496X}, and Santosh Kumar\orcidlink{0000-0003-2264-9014}}
  \IEEEauthorblockA{
    ashutoshs22102@iiitnr.edu.in, khushdeep22102@iiitnr.edu.in, amit22102@iiitnr.edu.in, \\ abhisheks@iiitnr.edu.in and santosh@iiitnr.edu.in\\
    Department of Data Science and Artificial Intelligence\\
   IIIT Naya Raipur, Chhattisgarh, India 493661
  }
}

\begin{document}
\maketitle
\begin{abstract}
In today's world, stress is a big problem that affects people's health and happiness. More and more people are feeling stressed out, which can lead to lots of health issues like breathing problems, feeling overwhelmed, heart attack, diabetes, etc. This work endeavors to forecast stress and non-stress occurrences among college students by applying various machine learning algorithms: Decision Trees, Random Forest, Support Vector Machines, AdaBoost, Naive Bayes, Logistic Regression, and K-nearest Neighbors. The primary objective of this work is to leverage a research study to predict and mitigate stress and non-stress based on the collected questionnaire dataset. We conducted a workshop with the primary goal of studying the stress levels found among the students. This workshop was attended by Approximately 843 students aged between 18 to 21 years old. A questionnaire was given to the students validated under the guidance of the experts from the All India Institute of  Medical Sciences (AIIMS)  Raipur, Chhattisgarh, India, on which our dataset is based. The survey consists of 28 questions, aiming to comprehensively understand the multidimensional aspects of stress, including emotional well-being, physical health, academic performance, relationships, and leisure. This work finds that Support Vector Machines have a maximum accuracy for Stress, reaching 95\%. The study contributes to a deeper understanding of stress determinants. It aims to improve college student's overall quality of life and academic success, addressing the multifaceted nature of stress.\end{abstract}

\vspace{0.1in}

\vspace{0.1in}
\section{Introduction}
\vspace{0.06in}
The college journey is an adventure of growth and self-discovery, challenging students to expand their minds and develop their character. It provides countless opportunities for personal and academic advancement, yet it also presents formidable obstacles that can profoundly impact students well-being. Recognizing and addressing these challenges is essential to creating a supportive and empowering college experience that meets the diverse needs of each individual \footnote{\url{https://www.usnews.com/education/best-colleges/articles/stress-in-college-students-what-to-know}}.

Stress detection encompasses various categories, including acute stress, chronic stress, episodic acute stress, eustress, and distress. Acute stress stems from short-term physiological changes triggered by specific events or pressures, while chronic stress persists due to ongoing conflicts or issues. Episodic acute stress manifests as repeated instances of acute stress, often affecting individuals with chaotic lifestyles. Understanding the complexities of stress, both positive and negative, is crucial as it affects not only individuals but society as a whole \cite{ahuja2019mental} and \cite{awada2023new}.

The emergence of the COVID-19 pandemic has brought into focus the widespread prevalence of stress and anxiety among students, as corroborated by recent research findings. A study conducted by The Center of Healing (TCOH) in India \cite{kumar2022novel} revealed a stark increase in stress and anxiety levels among a sample of over 10,000 respondents, underscoring the magnitude of the issue \cite{arsalan2022human}.

The two distinct ways of identifying stress are defined as (1) physiological and (2) psychological data. The major flaw in identifying stress by physiological data requires sensors to collect data under proper experimental conditions. This process can be both costly and time-consuming.  In this study, we propose a novel approach that utilizes psychological data for stress identification, leveraging its cost-efficiency and ease of implementation. By employing multiple machine learning algorithms and rigorous validation metrics such as accuracy, precision, recall, F1-score, and ROC curve analysis, our aim is to identify the most effective algorithm for stress classification. Our study utilizes real-time data collected from college students, providing valuable insights into stress detection and mitigation strategies tailored to the college environment.

This study offers a thorough exploration of stress detection among college students, shedding light on the challenges and opportunities in fostering student well-being within higher education. Through this study, we aspire to contribute to the development of impactful interventions and support systems aimed at bolstering the mental health and resilience of college students.

\vspace{0.1in}
\section{Related Work}
\vspace{0.06in}

To increase our understanding and the quality of our work, we have investigated various jobs in this field by many researchers, as discussed in this section.\\
In \cite{zhu2023stress} Lili Zhu et al. proposed a robust stress detection model integrating wrist-based electrodermal activity monitoring with machine learning algorithms. This innovative framework achieved a commendable accuracy rate of 92.90\% in discerning stress and non-stress states. However, it faces challenges in capturing subtle fluctuations in stress levels attributed to emotional dynamics. In \cite{nath2021smart} Nath et al. introduced a sophisticated deep learning-based model for stress detection utilizing wristband technology and cortisol as a stress biomarker, attaining an impressive accuracy rate of 94\%. Nonetheless, the practicality of widespread implementation may be hindered by the high cost and limited accessibility of the required devices.

Mahalakshmi et al. \cite{mahalakshmi2023predictions} delved into the realm of mental stress assessment among adolescents, employing a variety of machine learning algorithms. Their study highlighted the efficacy of the K-Nearest Neighbors model, which demonstrated a notable accuracy rate of 87.2\%. A significant contribution to stress detection research came in the form of the EmpathicSchool dataset proposed by Majid Hosseini et al. \cite{hosseini2022empathicschool}, offering a comprehensive collection of multimodal data, including facial expressions and physiological signals. This resource serves as a valuable asset for furthering our understanding of stress detection mechanisms.

The introduction of the Tr-Estimate technique by Rachakonda et al. \cite{rachakonda2022tr} for diagnosing Post-traumatic stress disorder (PTSD) represents a significant advancement in personalized care delivery. By leveraging continuous physiological signal monitoring and individualized medical history, this approach offers tailored interventions for PTSD patients. In the context of real-time stress prediction among students, Sinha et al. \cite{sinha2022educational} explored the efficacy of machine learning algorithms such as KNN and Naive Bayes, with Naive Bayes demonstrating promising results in swiftly identifying stress levels within educational environments. The integration of Internet of Things (IoT) technology and machine learning techniques, as demonstrated by Suhas et al. \cite{ks2022stress}, offers new avenues for investigating workplace stress. Notably, KNN emerged as the most effective algorithm in this context.

Aanchal Bisht et al. \cite{bisht2022stress} conducted a real-time database survey involving 190 school children, achieving accuracy 88\% 
rate using the K Nearest Neighbors algorithm. In a separate study, researchers utilized the HINTS database to extract 26 variables, employing a range of machine learning techniques to identify factors contributing to psychological distress \cite{barbayannis2022academic} Barbayannis et al. surveyed 843 college students, uncovering a notable correlation between high academic stress levels and poor mental well-being, particularly among first-year students \cite{chen2022machine}. These diverse research endeavors collectively contribute to the ongoing efforts in stress detection and mitigation, providing valuable insights and paving the way for future exploration in this critical domain.

\vspace{0.05in}
\subsection{Problem Statement}
\vspace{0.05in}
College students encounter various social and economic challenges that significantly impact their academic performance and well-being. Academic pressure, financial constraints, social isolation, and personal issues are among the myriad factors contributing to their stress levels. This stress not only affects their mental health and academic achievements but also has broader implications for their overall well-being. Traditional statistical methods often fall short of comprehensively understanding the intricate relationships between these stress predictors identified in previous research. Therefore, there is a pressing need to employ advanced machine learning algorithms to accurately predict stress levels among college students. Early detection of stress is imperative, as evidenced by previous studies, and sensor data could be instrumental in this regard. However, deploying sensors across various locations poses challenges due to the associated costs and time constraints, particularly given the scale of the survey.

\vspace{0.05in}
\subsection{Aim of this Study}
\begin{enumerate}
\item Conduct research and implement stress measurement procedures.
\item Gather current data through surveys.
\item Enhance understanding of stress factors for preventive measures.
\item Improve students quality of life and academic success by addressing stress.
\end{enumerate}

\vspace{0.1in}
\section{Methodology}
\vspace{0.06in}

\subsection{Proposed Architecture}
\begin{figure}[ht]
    \centering
\includegraphics[width=\linewidth]{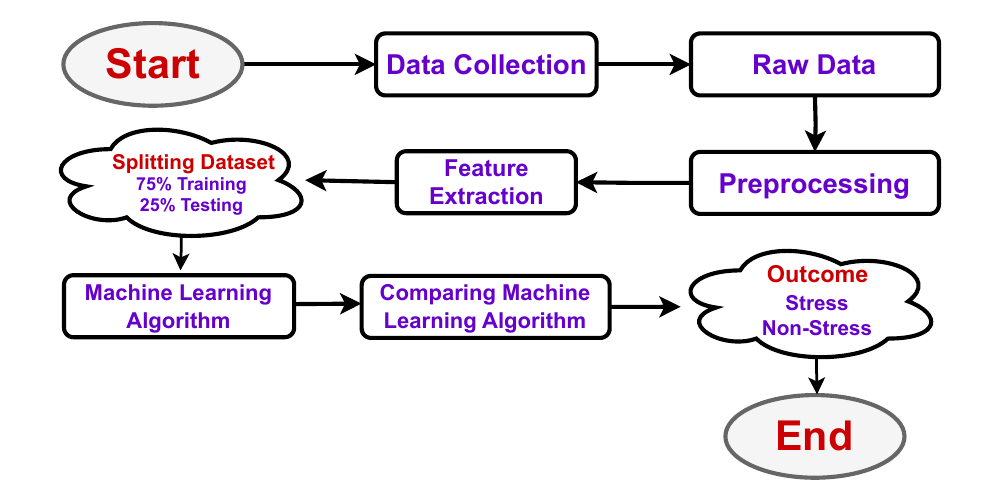}
    \caption{Overall Architecture of the Proposed Work}
    \label{Fig.1}
\end{figure}

The functionality of the proposed framework is depicted in (Fig-\ref{Fig.1}). Comprising several modules, the framework includes (1) Data collection, (2) Preprocessing of a raw dataset, (3) Feature extraction, (4) Splitting the dataset into training and testing, (5) Applying machine learning algorithm, (6) Comparing the performance of machine learning algorithm, and (7) Result in stress and non-stress. Further details regarding each module are provided in the subsequent subsections.

\vspace{0.05in}
\subsection{Dataset description}
\begin{figure}[h]
    \centering
    \includegraphics[width=\linewidth]{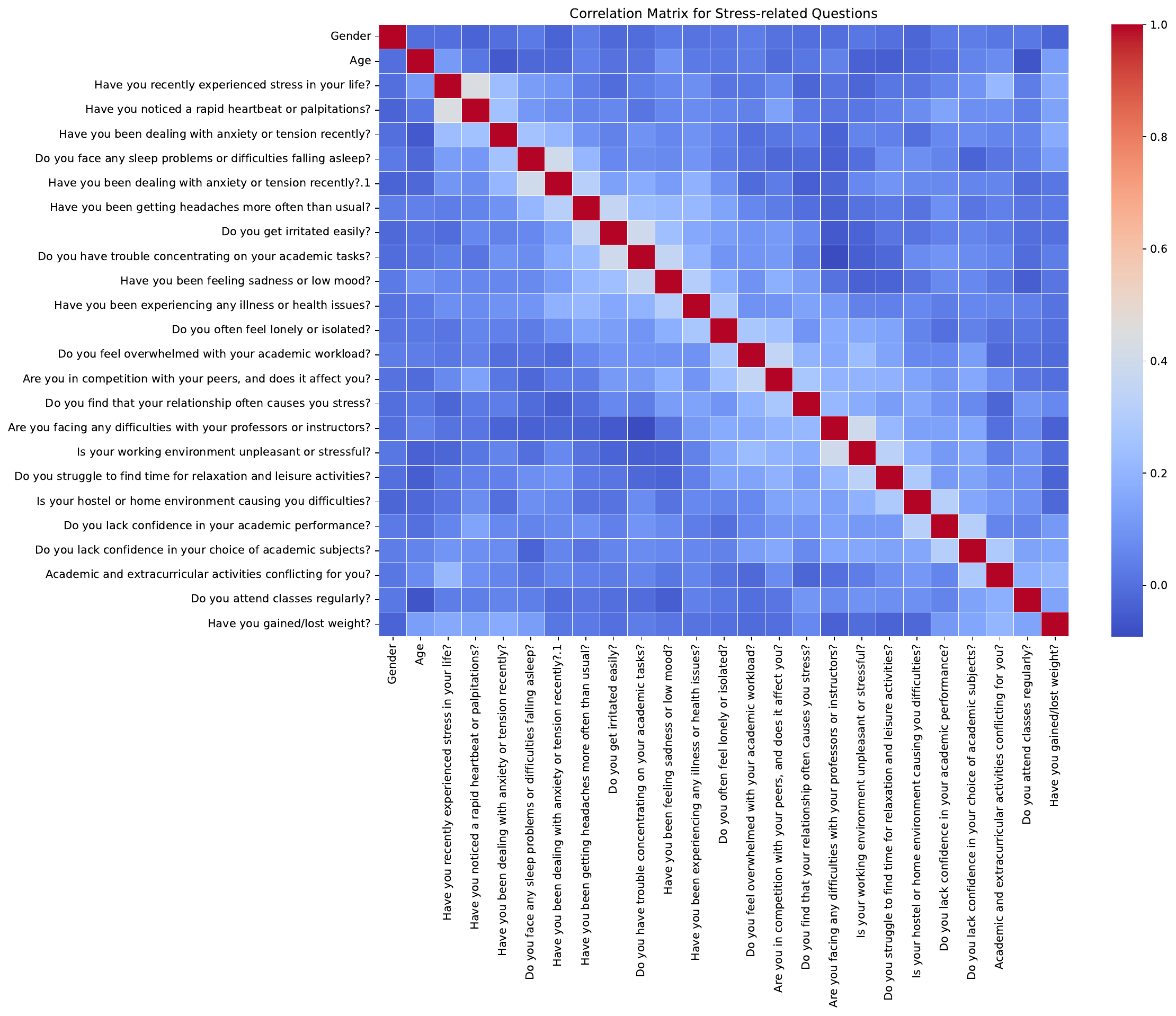}
    \caption{Heatmap of the Dataset Attributes}
    \label{Fig.2}
\end{figure}

We conducted a workshop under expert guidance to assess students stress levels. Various college students participated, randomly receiving stress-related questionnaires. Real-time data was gathered from 843 college students, comprising 548 males and 295 females, aged between 18 to 21 years, using a Google Form. The resulting dataset is organized in CSV format and includes 28 attributes for statistical analysis of stress levels reported by the students.\\ These attributes correspond to a series of questions posed to college students, addressing their experiences and feelings concerning stress over the past two months. The questions are categorized into seven distinct groups: Stress and Emotional Well-being, Physical Well-being, Academic Performance, Relationships and Social Environment, Leisure and Relaxation, and Stress and Non-Stress levels. Additionally, we collected the necessary personal details from the participants. The questions offer five response options ranging from "Not at all" to "Extremely." These categories are used to label and organize the collected data, as depicted in (Fig-\ref{Fig.2}), providing a comprehensive overview of the students stress-related experiences across various dimensions. Dataset available on kaggle\footnote{\url{https://www.kaggle.com/datasets/ashutoshsingh22102/stress-and-well-being-data-of-college-students}}

\vspace{0.02in}
\subsection{Preprocessing of collected dataset}
\begin{figure}[ht]
    \centering
    \includegraphics[width=\linewidth]{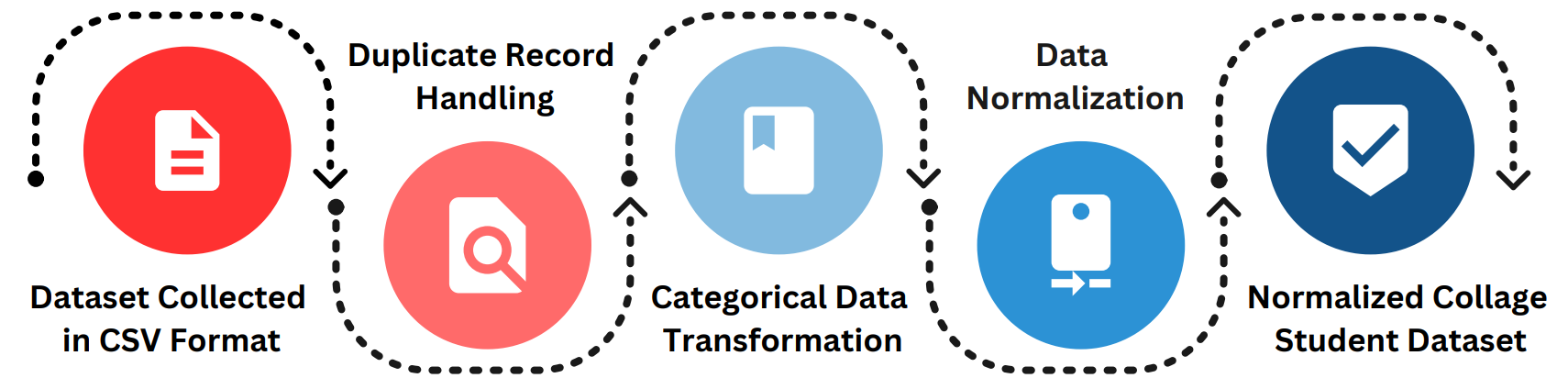}
    \caption{Architecture of Data Preprocessing Workflow}
    \label{Fig.3}
\end{figure}

\vspace{0.02in}
Addressing issues in data preprocessing stands as one of the most critical tasks during data processing. This process involves employing various techniques to tackle missing values, anomalies, and inconsistencies. As the significance of data grows across research, business, and academia, the need for more advanced analysis methods becomes paramount, particularly with larger datasets. In our model, we have implemented several existing data preprocessing techniques, including duplicate handling, encoding, and normalization, as illustrated in (Fig-\ref{Fig.3}). \\ In this study, the initial step involves data collection. Following data collection, the dataset often contains numerous duplicate records that require handling. Subsequently, the normalization process is applied. This leads to the finalization of the normalized college dataset.

\vspace{0.05in}
\subsection{Algorithms}
\vspace{0.02in}
Classification algorithms act as intelligent tools that efficiently organize information into appropriate categories with minimal errors. In this study, we employed seven machine learning algorithms to assess stress levels among college students. Initially, we trained our model using 75\% of the data to familiarize it with stress patterns. Subsequently, we evaluated its accuracy by testing it on the remaining 25\% of the data.

\vspace{0.05in}
\subsubsection{Decision Trees}
Decision trees neatly organize data by features, forming a hierarchical structure for classification. We utilized decision trees to categorize stress levels, with each node assessing feature importance. This method provided detailed insights into the hierarchical significance of factors affecting stress determination.

\vspace{0.02in}
\subsubsection{Random Forest}
In this, we applied Random Forest, an ensemble learning method, to classify stress levels. By independently constructing multiple trees and combining their predictions, we achieved enhanced accuracy in assessing attribute importance for determining stress severity in students.

\vspace{0.02in}
\subsubsection{Support Vector Machine}
Support Vector Machines (SVMs) efficiently classify stress severity by identifying optimal hyperplanes. In this work, SVMs crucially distinguish stress classes by leveraging support vectors for precision enhancement, contributing to accurate stress level discernment in our classification framework.

\vspace{0.02in}
\subsubsection{AdaBoost}
AdaBoost iteratively enhances classification accuracy by focusing on misclassified instances. In our study, AdaBoost significantly improved the performance of our stress level classification model by combining weak learners into a robust classifier. Its ability to identify patterns in stress data aids in accurate stress level prediction.

\vspace{0.02in}
\subsubsection{Naive Bayes}
Naive Bayes, applying the Bayes theorem, assumes feature independence and categorizes stress levels in this work. It evaluates the likelihood of instances falling into classes based on features, efficiently capturing stress factor relationships. Its simplicity enhances our research's thorough analysis of stress factors.

\vspace{0.02in}
\subsubsection{Logistic Regression}
Logistic Regression, a regression analysis technique, was employed to predict stress severity, focusing on binary classification. It models the probability of the default class, capturing the relationship between input features and class probability, contributing to precise stress level classification.

\vspace{0.02in}
\subsubsection{k-Nearest Neighbors}
We utilized the k-Nearest Neighbors algorithm for stress severity prediction, leveraging the similarity of neighboring data points. By determining the majority class among the k-nearest neighbors, the algorithm accurately predicts stress classification based on the support of neighboring instances.

\vspace{0.05in}
\subsection{Statistical Measures}
To assess stress conditions accurately, we employed statistical measures to evaluate the effectiveness of our model. These measures are crucial for analyzing the model's performance comprehensively.

\begin{equation}
    \text{Precision} = \frac{TP}{TP + FP}
\end{equation}

\begin{equation}
    \text{Recall} = \frac{TP}{TP + FN}
\end{equation}

\begin{equation}
    \text{F1-score} = 2 \times \frac{\text{Precision} \times \text{Recall}}{\text{Precision + Recall}}
\end{equation}

\begin{equation}
    \text{Accuracy} = \frac{TP + TN}{TP + TN + FP + FN}
\end{equation}

\vspace{0.5cm}
\section{Results and Discussions}

\vspace{0.05in}
\subsection{Performance Analysis}
\begin{figure}[ht]
    \centering
    \includegraphics[width=\linewidth]{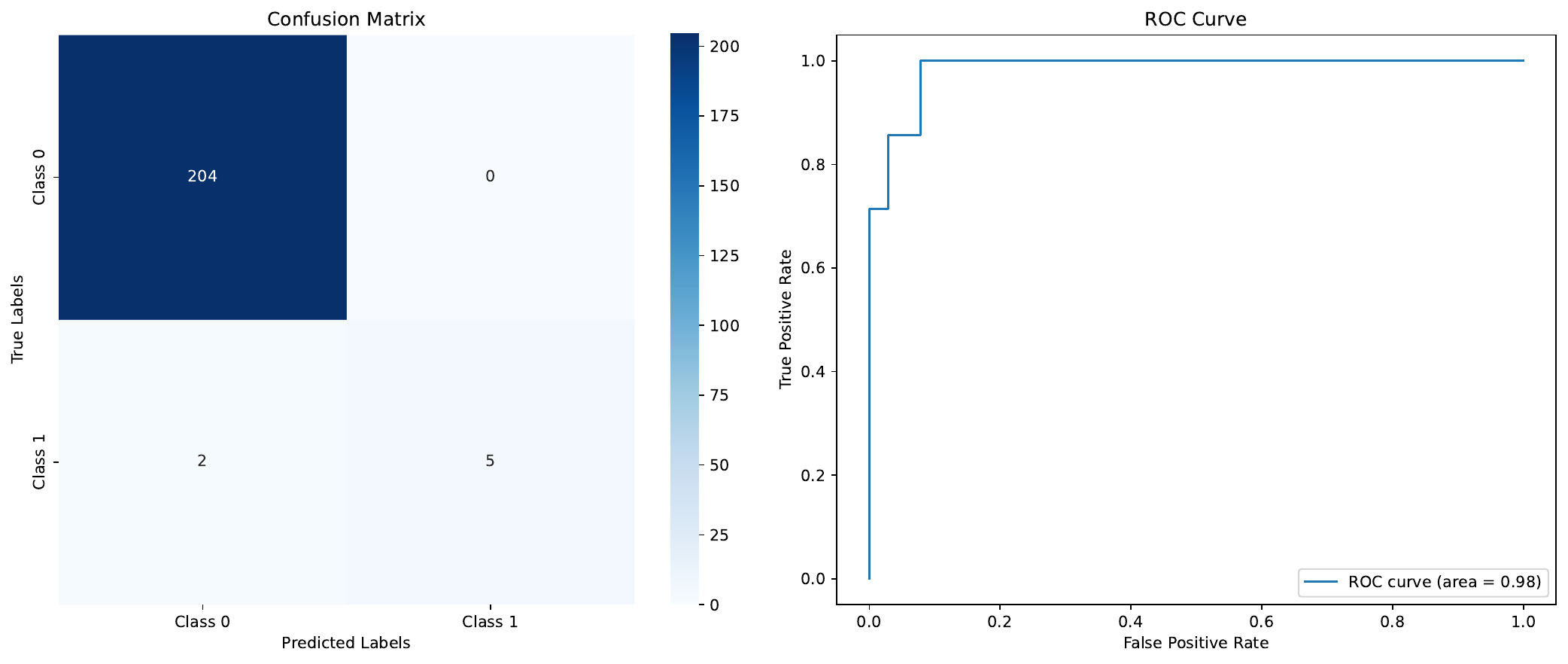}
    \caption{Confusion Matrix (Left) and ROC Curve (Right) for Stress  Classification}
  \label{Fig.4}
\end{figure}

The performance assessment of the proposed framework is conducted utilizing 5-cross-validation benchmark settings, as illustrated in (Fig-\ref{Fig.4}) referring to SVM which has the highest accuracy in this work. This analysis presents a comprehensive evaluation of stress severity classification through statistical measures. The results offer an insightful overview of stress classification outcomes, leveraging the confusion matrix and ROC curve. The delineated confusion matrix and ROC curve demonstrate precise stress measurement classification, thereby emphasizing the effectiveness of the classification model .

\vspace{0.05in}
\subsection{Comparison Between Different Machine Learning algorithms}
\begin{figure}[ht]
    \centering
    \includegraphics[width=0.4\textwidth]{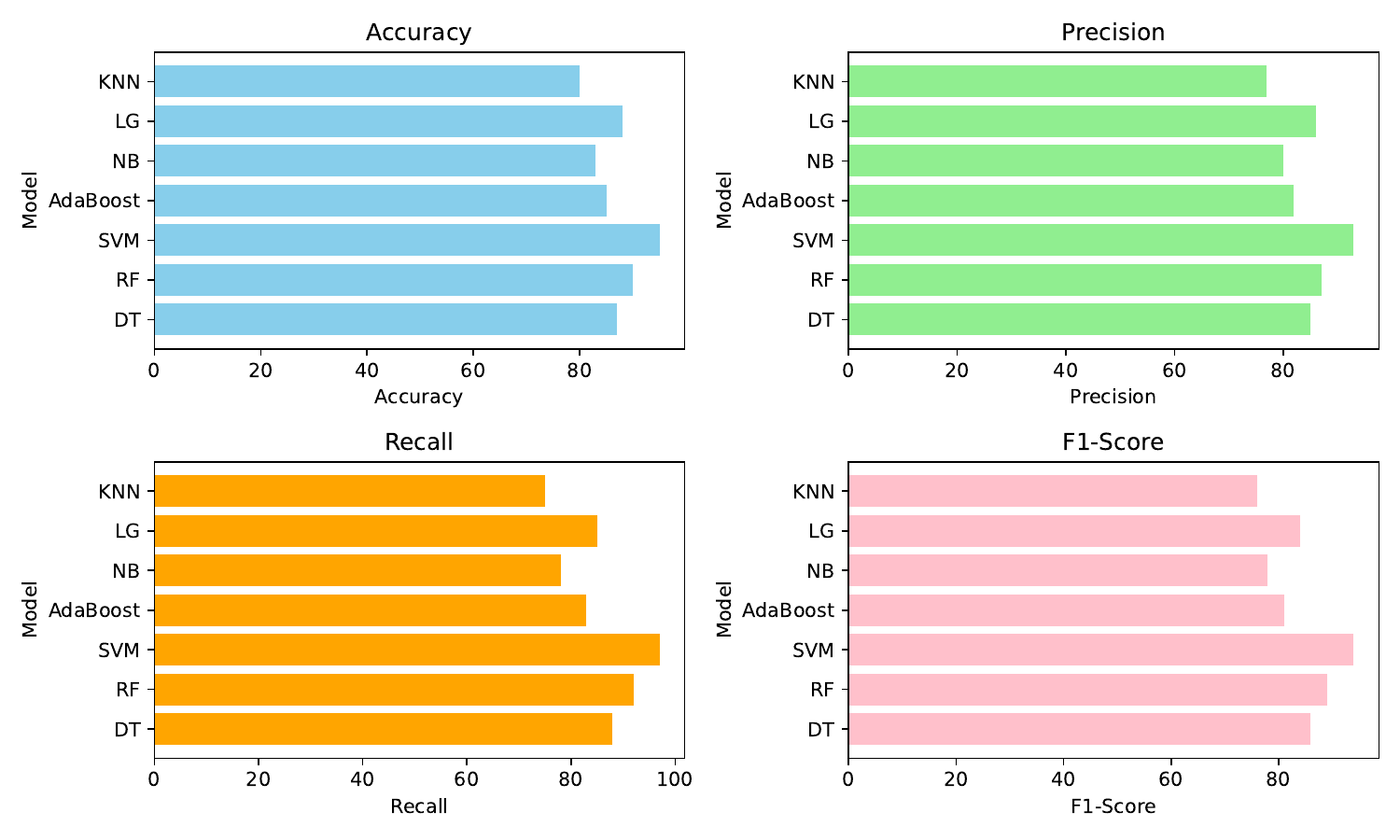}
    \caption{Comparing several machine learning models for Stress  Classification}
   \label{Fig.5}
\end{figure}

\vspace{0.02in}
 We conducted a comprehensive comparison of various machine-learning algorithms for stress classification. We compared the performance of the proposed system with state-of-the-art methods. These encompassed Decision Trees (DT), Random Forest (RF), Support Vector Machines (SVM), Adaptive Boosting (AdaBoost), Naive Bayes (NB), Logistic Regression (LR), and K-nearest Neighbors (k-NN). The (Fig-\ref{Fig.5}) provides a graphical representation of the accuracy, precision, recall, and F1 score for each algorithm, facilitating a detailed analysis of their efficacy in stress classification. Additionally, (Table-1) presents the numerical values of these metrics, revealing that Support Vector Machines achieved maximum accuracy for Stress at 95\%. This study underscores the potential of our approach in deciphering college students' sentiments and their decision-making in college activities.

\begin{table}[!htb]\small
 \begin{center}
 \caption{Comparison of Different Performance Algorithms for Stress  Classification}
\begin{tabular}
{|c|c|c|c|c|}\hline
\textbf{Model} &\textbf{Accuracy}&\textbf{Precision}&\textbf{Recall}&\textbf{F1-Score}\\\hline 
      DT & 87 & 85 & 88 & 86\\
      \hline
      RF & 90 & 87 & 92 & 89\\
      \hline
      SVM & 95 & 93 & 97 & 94\\
      \hline
      AdaBoost & 85 & 82 & 83 & 81\\
      \hline
      NB & 83 & 80 & 78 & 78\\
      \hline
      LG & 88 & 86 & 85 & 84\\
      \hline
      KNN & 80 & 77 & 75 & 76\\
      \hline
\end{tabular}
\end{center}
\label{Table.1}
\end{table}

\section{Conclusion and Future Direction}
This study highlights the efficacy of utilizing a questionnaire-based approach for initial stress analysis among individuals. By applying seven classification algorithms to data collected from 843 students from Chhattisgarh, India, we found Support Vector Machines to exhibit the highest accuracy in stress classification, reaching 95\%. This underscores the potential of machine learning techniques in aiding individuals in understanding and managing their stress levels effectively.

In the future, our work will focus on two critical avenues for advancement. Firstly, we aim to expand the size of our stress dataset to encompass a broader spectrum of stress conditions, enabling a more comprehensive analysis and refined classification accuracy. Additionally, we plan to explore the integration of multimodal sensor data from wrist-worn wearable devices for stress detection. By employing advanced deep learning techniques, we aim to enhance the accuracy and granularity of stress classification, ultimately contributing to more personalized and effective stress management strategies.

\section*{Data and Code Avialbility} 

We provide open course code via Github repository \footnote{\url{https://github.com/ashutosh22102/Machine-Learning-Algorithms-for-Detecting-Mental-Stress-in-College-Students}}

\bibliographystyle{plain}  
\bibliography{cas-refs} 
\end{document}